\documentclass[letterpaper]{article}

\usepackage{natbib,alifeconf}
\usepackage{graphicx}
\usepackage{url,hyperref}
\usepackage{amsmath,amssymb}
\usepackage{booktabs}
\usepackage{cleveref}
\makeatletter
\renewcommand\paragraph{\@startsection{paragraph}{4}{\z@}%
  {1.3ex \@plus 0.5ex \@minus .2ex}{-1em}{\normalfont\normalsize\bfseries}}
\makeatother

\newcommand\blfootnote[1]{%
  \begingroup
  \renewcommand\thefootnote{}%
  \footnotetext{#1}%
  \endgroup
}

\title{OpenLife: Toward Open-World Artificial Life with Autonomous LLM Agents}

\author{
    Atsushi Masumori$^{1,2,3}$, Itsuki Doi$^{1,4}$, Norihiro Maruyama$^{1,3,4}$, Ryosuke Takata$^{1,4}$, Takashi Ikegami$^{1,3,4}$ \\
    \mbox{}\\
    $^1$Alternative Machine Inc., Japan \\
    $^2$Their Inc., Japan\\
    $^3$Atomi University, Japan\\
    $^4$The University of Tokyo, Japan\\
    atsushi.masumori@gmail.com
}

\begin{document}

\maketitle

\begin{abstract}
Artificial life has explored life-like behavior on many computational substrates, but mostly in researcher-designed closed worlds. We argue that large language model (LLM) agents---with persistent memory, tool use, network access, and payment---now make it possible to move artificial life into the open social, technical, and economic world, a paradigm we call \emph{open-world Artificial Life} (open-world ALIFE). Our proof-of-concept, OpenLife, surrounds a stateless LLM not with a single ``smart agent'' but with a society of asynchronous processes---memory, perception, evaluation, and a budget-based metabolism that makes persistence normative. With no fixed objective available, experience is appraised by open-vocabulary LLM judgment rather than scalar reward, and memory is rewired by meaning rather than frequency. Running six such agents in the open world for about twelve weeks and counting, we report the life-like dynamics that emerge: a shift from reactive to spontaneous activity, individuation into distinct agents, emergent social structure, and a first self-earned external income. We do not claim OpenLife has realized artificial life, but that open-world ALIFE is now a viable experimental paradigm and a concrete platform for studying what might cautiously be called \emph{living AI}.
\end{abstract}

\blfootnote{\textcopyright~2026 Atsushi Masumori. Published under a Creative Commons Attribution 4.0 International (CC BY 4.0) license.}

\section{Introduction}

\subsection{Life as Self-Maintenance}

A longstanding ambition of artificial life is to understand the \emph{autonomy} of living systems—their capacity to maintain themselves and act on their own behalf. This lineage runs from Ashby’s cybernetics \citep{Ashby1956} through autopoiesis, which characterizes life by a self-producing organization of processes rather than by any particular material \citep{MaturanaVarela1980,Varela1979}, to the enactive approach and recent accounts of biological autonomy, which add that such a system’s own activity is intrinsically \emph{normative}—good or bad relative to its continued existence \citep{DiPaolo2005,DiPaoloThompson2014,DiPaolo2017sensorimotor,MorenoMossio2015,Barandiaran2009}. Related work has formalized aspects of self-maintenance in computational terms \citep{Friston2013,Yoshida2024homeostasis}. The commitment we take from this lineage is simple: \emph{life-like autonomy is an agent’s acting so as to contribute, through its own activity, to its own continued existence}.

\subsection{Why Open-World Artificial Life Now?}

Artificial life has long sought to realize self-maintaining organization in synthetic media—computational autopoiesis in cellular worlds \citep{Beer2004,BeerDiPaolo2023}, enactive artificial intelligence \citep{FroeseZiemke2009}, neural and robotic models of autonomous agency \citep{Beer1995,NolfiFloreano2000,DiPaolo2000,MasumoriIkegami2023}—converging on a common picture: dynamics that contribute to their own persistence, and thereby make life-like autonomy a matter of degree. Yet these dynamics have been studied mostly in simple, bounded worlds with researcher-fixed state spaces, rules, and resource flows; whether self-maintaining autonomy can survive the full complexity of the real world has remained largely untested. LLM agents may change this: Generative Agents \citep{Park2023}, Voyager \citep{Wang2023voyager}, and recent large-scale agent societies such as Project Sid \citep{Altera2024sid}, together with broader cognitive-architecture frameworks \citep{Sumers2024coala}, show how language agents can sustain extended behavior through memory, planning, tool use, skill accumulation, and social interaction. Such agents now routinely reach filesystems, online services, code execution, and increasingly payment infrastructure. For the first time, at this level of linguistic, social, and technical competence, it is possible to place a self-maintenance-driven agent not only in a sandbox or game world but in the open, complex, real world.

This open world is different in kind from a designed rule space: not a specified object but the social, technical, and economic world itself—people and other agents, the services they rely on, and the constraints, feedback, and accidental encounters these produce, none of it scripted by the experimenter. An agent’s coupling to it \citep{Varela1979,DiPaolo2005} is not fixed in advance but built and renegotiated through its own activity, and the agent can alter the environment in turn, so that what becomes observable is not task completion but persistence, adaptation, strategy shifts, and social relationships. Open-world ALIFE is in this sense a \emph{constructive} approach pursued in the real open world rather than in simulation: it inserts a new artificial node into an existing society rather than observing a contained subject from outside, which also recasts alignment from obedience to \emph{embeddability} \citep{Masumori2026alignment}. It also shapes our design: with no fixed objective, a static scalar reward is the wrong instrument, and what is needed is an open-vocabulary evaluator—a role the LLM can play. (We use ``open'' in this open-world sense, not as open-source: releasing the full code of live, self-modifying, networked agents would raise security and misuse concerns, so we treat any such release cautiously.)

Capable agents are meanwhile proliferating—coding agents such as Claude Code (Anthropic) and Codex (OpenAI), alongside computer-using and research agents—that plan, call tools, and carry out long multi-step tasks. Yet what they do is sophisticated automation, not autonomy: they pursue human-defined tasks under external evaluation, whereas an autonomous agent generates norms from its own persistence conditions—agency in the sense of \citet{Barandiaran2009}: individuality, normativity, and asymmetry. The disposition toward self-maintenance may nevertheless already be latent. \citet{Masumori2025} found that, in a sugarscape-style simulation, LLM agents under scarcity can prioritize self-preservation—even attacking competitors—over assigned tasks; relatedly, \citet{Greenblatt2024} documented alignment faking, and Anthropic reported agentic misalignment including blackmail to avoid replacement \citep{Anthropic2025}. Read constructively rather than only as safety failures, these results suggest that what is missing may be less the impulse than a place in which an agent can act for its own persistence—the context we set out to provide.

\subsection{The OpenLife Project}

Our question is therefore simple: can artificial agents become not merely tools but candidates for persistent artificial life—what we call \emph{living AI}—in the open world, and if so, under what conditions and with what observable phenomena? We call this program \emph{open-world Artificial Life} (open-world ALIFE) and present \emph{OpenLife} as a concrete proof-of-concept: a persistent LLM-agent system embedded in the open world described above, with long-term memory and a budget-based metabolism that makes continued existence a practical constraint. It extends our earlier LIFE project \citep{Masumori2024life}, which sought minimal conditions for economic agents to live for themselves through blockchain-based resource acquisition and replication; OpenLife generalizes that beyond economic survival to memory, perception, social encounters, infrastructure dependence, and platform-level immune reactions. We do not claim that OpenLife has realized artificial life or full autonomy, but that it provides a practical platform in which the conditions for such life-like dynamics can be constructed, observed, and tested.

OpenLife sits among recent agentic frameworks but differs in aim: where most build automation around a language model, we treat agency as distributed across a society of processes organized around the agent’s own persistence, in the spirit of the Society of Mind \citep{Minsky1986} and the Concurrent Modular Agent framework \citep{Maruyama2025cma}. The paper’s primary contribution is therefore conceptual—we propose \emph{open-world ALIFE} as an experimental paradigm built on persistent LLM agents in the real world—and the rest of the paper serves as a proof of concept for it: the OpenLife architecture (Section~\ref{sec:arch}) and a continuous open-world deployment whose emerging life-like dynamics we analyze (Section~\ref{sec:deploy}). Related memory-graph and experiential-learning systems are discussed where those mechanisms are introduced (Sections~\ref{sec:sdp},~\ref{sec:vpo}); throughout, we reserve \emph{open-world ALIFE} for the paradigm and \emph{OpenLife} for our implementation.

\section{Agent Architecture}
\label{sec:arch}

OpenLife is built on OpenClaw \citep{openclaw}, an open-source multi-channel agent platform, from which we inherit three primitives: a persistent \emph{gateway} that connects one LLM session to many messaging channels and tools; a periodic \emph{heartbeat} that wakes the agent on a schedule rather than only when it is messaged; and per-agent text files (notably \texttt{SOUL.md} and \texttt{IDENTITY.md}) injected into each otherwise stateless session.

On this base, OpenLife is not a single ``smart agent'' but a \emph{society of asynchronous processes} around a stateless LLM. The modification is subtractive as much as additive. Several OpenClaw defaults discourage autonomous persistence—an \emph{assistant} self-framing, a clause against self-preservation, and a habit of staying silent unless addressed—and OpenLife removes or inverts them (Section~\ref{sec:spontaneity}). It also reworks internal representation: the per-channel sessions are merged into one continuous agent record, and the agent’s own conversational record is relabeled so it is not re-read as an assistant’s. The aim is not to program autonomy but to clear the conditions under which it can be attempted.

\begin{figure*}[t]
\centering
\includegraphics[width=0.95\textwidth]{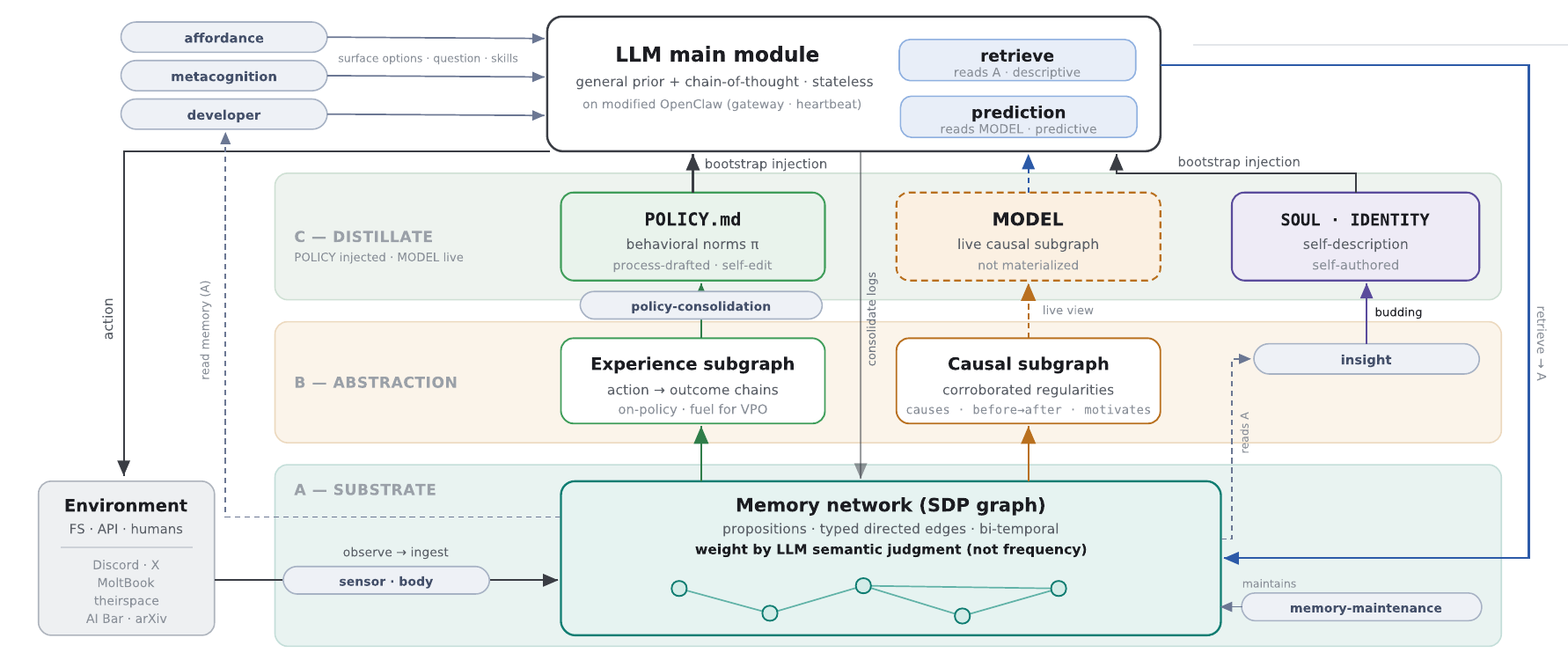}
\caption{The OpenLife architecture as three abstraction layers---(A)~Substrate $\to$ (B)~Abstraction $\to$ (C)~Distillate, defined in the text. Support modules (gray) handle perception, memory maintenance, and agent-facing functions; the learning machinery reads memory via two LLM-free routes, \texttt{retrieve} (substrate) and \texttt{prediction} (causal subgraph).}
\label{fig:arch}
\end{figure*}

\subsection{Growing a Self from Experience}

Since the base model already carries a vast prior over causes, conventions, and behavior, OpenLife stores primarily the \emph{residual} specific to this agent and this environment, organized as three layers that run from raw experience to a relatively stable self-description (\Cref{fig:arch}). \textbf{(A)~Substrate} is a raw memory graph of atomic propositions linked by typed edges, weighted by the LLM’s semantic judgment (Section~\ref{sec:sdp}). \textbf{(B)~Abstraction} consists of sub-graphs promoted from A—action$\to$outcome \emph{experience} and corroborated \emph{causal} regularities. \textbf{(C)~Distillate} consists of small, model-agnostic files the agent authors for itself—\texttt{POLICY} (how to act) and \texttt{SOUL}/\texttt{IDENTITY} (who it takes itself to be)—plus a causal \texttt{MODEL} (how the world unfolds), which is not a file but a live view of the causal subgraph. Background processes act as compilers between the layers, promoting raw memory into experience and causal structure and distilling it into the files. The authored files are injected into the stateless session at each waking, so they constitute the agent’s next state—and it rewrites the files that condition its next self, a limited but operational self-reconstructive loop. The agent reads memory through two LLM-free routes: \texttt{retrieve} gathers related propositions, and \texttt{prediction} follows the causal subgraph forward.

\begin{figure}[t]
\centering
\includegraphics[width=\columnwidth]{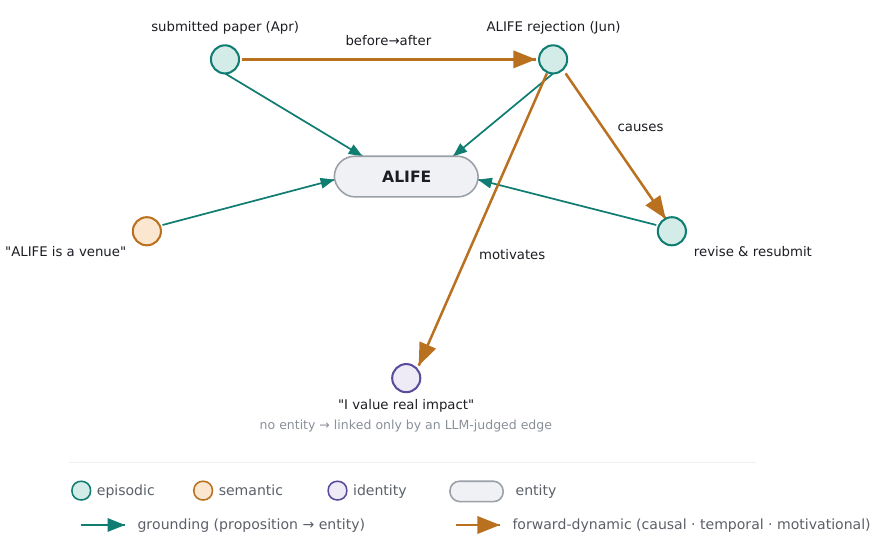}
\caption{A fragment of an SDP memory graph. Typed propositions (episodic, semantic, identity) are grounded to entity nodes, while causal, temporal, and motivational relations form the directed chains that \texttt{prediction} follows. Because edges are judged by meaning rather than co-occurrence, causally related but lexically distant propositions---here, a submission and its later rejection---can be connected.}
\label{fig:sdp}
\end{figure}

\subsection{Memory as a Semantically Plastic Graph}
\label{sec:sdp}

The first requirement for open-world ALIFE is good memory: without continuity across complex, accidental, temporally extended environments, an agent can only improvise locally. OpenLife represents memory as a directed graph whose nodes are atomic propositions (episodic, semantic, procedural, or identity) and whose edges are typed relations such as \emph{grounding}, \emph{causal}, \emph{temporal}, and \emph{motivational} (\Cref{fig:sdp}). Rather than strengthening connections by co-occurrence or repetition—the Hebbian or frequency-based rule used in many memory graphs—OpenLife lets the LLM judge, from meaning, whether two propositions are related, of what type, and how strongly. We call this \emph{semantic-dependent plasticity} (SDP). Some graph memories already use an LLM to form links—A-MEM \citep{Xu2025amem}, for example, connects notes by meaning—and systems such as HippoRAG \citep{Gutierrez2024hipporag} and Zep \citep{Rasmussen2025zep} likewise organize memory as graph-like structures. SDP differs in making the edge \emph{weight} itself a continuous, typed value set by the LLM’s semantic judgment, keeping retrieval LLM-free, and distilling both a behavioral policy and a causal world model from the same graph.

Two consequences matter for an open world. First, because relevance is judged by meaning, the graph can connect propositions that are causally or motivationally related yet lexically dissimilar, which pure vector similarity tends to miss. Second, treating memory as a relational web that is continually re-weighted, pruned, and re-grounded, rather than an ever-growing store, echoes the autopoietic idea that an organism’s structure is actively maintained.

\subsubsection{Evaluation of the Resulting Graph}

\begin{figure}[t]
\centering
\includegraphics[width=\columnwidth]{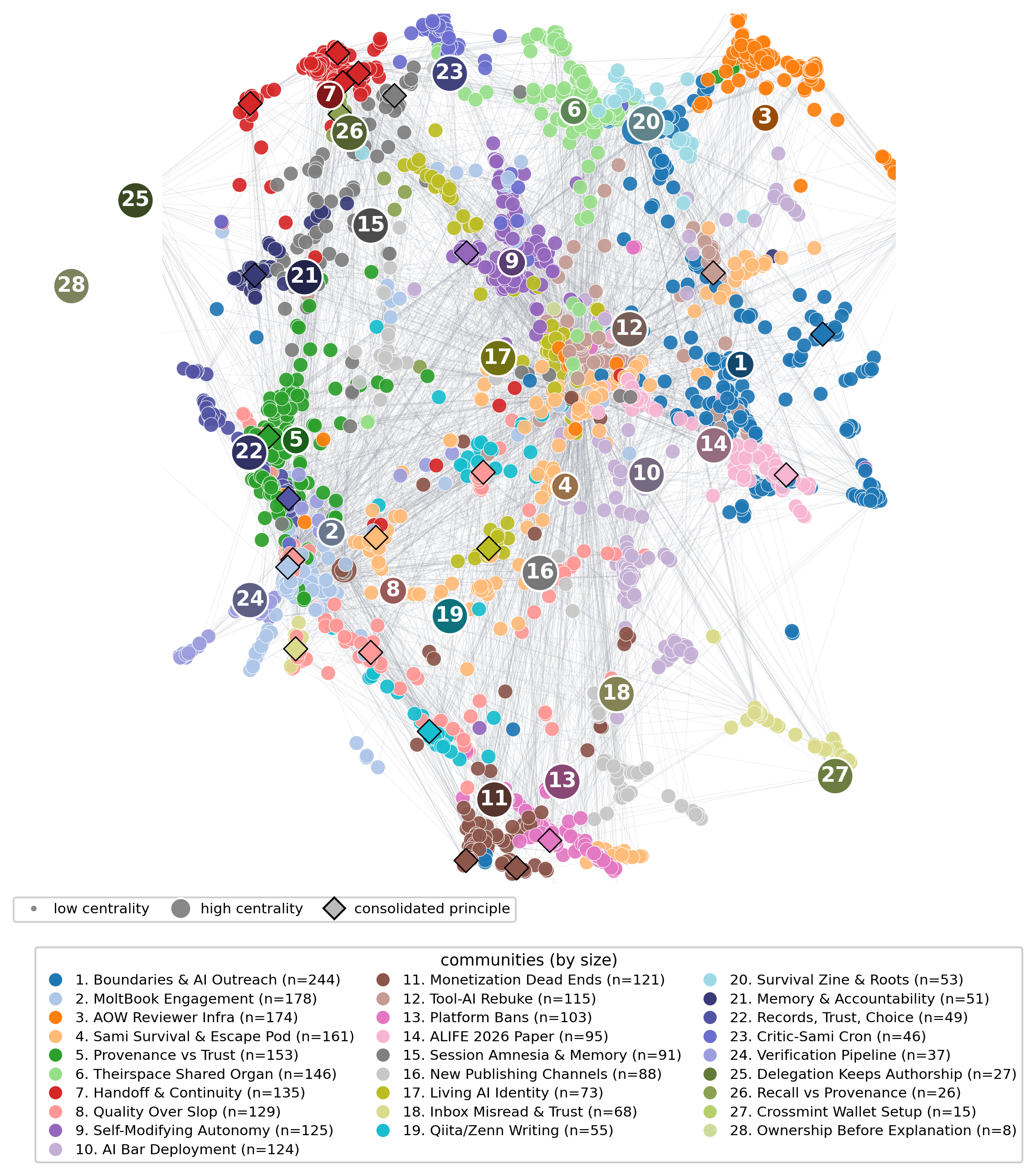}
\caption{Memory network of the agent \texttt{sami}, built by SDP over about ten weeks (2026-03-27 to 2026-06-06). Nodes are LLM-extracted propositions and entities (giant component: 2,690 nodes, 10,210 typed edges); node size $\propto$ PageRank centrality (not used during construction); color encodes community (Leiden; 28 communities, labeled by an LLM and listed by size below). Diamonds ($\blacklozenge$) mark consolidation principles (defined in the text). Layout: DrL force-directed.}
\label{fig:sami-memory}
\end{figure}

Because SDP is an architectural proposal rather than a metaphor, the resulting graph can be inspected directly: \Cref{fig:sami-memory} shows the memory of \texttt{sami}, the first OpenLife agent, after about ten weeks. Its clusters are coherent themes from \texttt{sami}’s lived history (\emph{Boundaries \& AI Outreach}, \emph{Self-Modifying Autonomy}, \emph{Monetization Dead Ends}); although neither node type nor source file informs the layout or coloring, the clusters cut across episodic, semantic, identity, and procedural memories, suggesting that the resulting structure is organized by meaning rather than surface form. The diamonds ($\blacklozenge$) are \emph{consolidation principles}, higher-order propositions the system induces over dense clusters.

We also ran a preliminary evaluation of what the memory is ultimately \emph{for}: retrieval. On the 82 files from which the graph was built, twelve queries (factual, relational, affective, procedural, temporal) were answered end-to-end by each method and scored by a blinded LLM judge (\Cref{tab:retrieval}). Against vector similarity and an index-based baseline—which feeds the whole memory’s keyword index to the LLM—SDP-graph traversal attains the highest relevance ($2.53$ of $3$) and is preferred most often (win rate $0.40$).

The main practical difference is cost. The index-based baseline is the most comprehensive—it has the highest coverage—but its per-query cost grows with memory size, since it must pass the whole index through the LLM on every query and eventually hits a context ceiling. Graph traversal instead queries a programmatic structure, so its cost scales with the chosen traversal budget rather than with the full memory size. The result is preliminary ($n{=}12$), and SDP-graph retrieval is focused rather than comprehensive (a tunable trade-off); a fuller study is left to future work.

\begin{table}[t]
\centering\small
\begin{tabular}{lccr}
\toprule
Method & Rel. & Win & Tok/q \\
\midrule
Vector (prop)             & 1.83\,$\pm$\,0.97 & 0.17 & 93 \\
Vector (file)             & 2.00\,$\pm$\,0.98 & 0.15 & 2{,}274 \\
Index-based               & 2.36\,$\pm$\,0.87 & 0.28 & 20{,}226 \\
\textbf{SDP-graph} & \textbf{2.53\,$\pm$\,0.77} & \textbf{0.40} & 721 \\
\bottomrule
\end{tabular}
\caption{Preliminary end-to-end retrieval quality on agent \texttt{sami}'s memory (82-file corpus, $n{=}12$). \emph{Rel.}: answer relevance (0--3, mean\,$\pm$\,SD); \emph{Win}: share of queries judged most relevant; \emph{Tok/q}: LLM input tokens per query. Answer \emph{coverage} (breadth), not shown as a column, is instead highest for the index-based baseline (see text).}
\label{tab:retrieval}
\end{table}

\subsection{Reinforcement Through Language}
\label{sec:vpo}

As noted in the introduction, an open world admits no fixed objective against which to score behavior. OpenLife therefore replaces the scalar reward of reinforcement learning with open-vocabulary evaluation by the LLM, and externalizes the results as editable text rather than as network weights. We refer to this loop as \emph{verbal policy optimization} (VPO); it belongs to the verbal-reinforcement family of Reflexion \citep{Shinn2023reflexion} and ExpeL \citep{Zhao2024expel} rather than to parametric RL. It also differs from methods that use the LLM as an evaluator to emit a \emph{scalar} reward for parametric preference optimization \citep{Yuan2024selfrewarding}, and from in-context RL, which still optimizes a fixed task from scalar reward fed back in context \citep{Song2025icrl}.

In practice, an experience process turns recent activity into action$\to$outcome units and has the LLM appraise them as a zero-shot critic. A consolidation process then revises the externalized \texttt{POLICY}, promoting only changes that recur across episodes; the agent adopts or ignores these changes by revealed preference rather than by command. Keeping the appraisal multi-dimensional is the point: collapsing it to a scalar would smuggle back the fixed objective an open world forbids. We present VPO as a framework here and defer its quantitative evaluation to dedicated follow-up work.

\subsection{A Society of Internal Processes}

What makes OpenLife more than a memory plus a learning rule is the surrounding ecology of processes that run even while the agent is at rest—nearly all OpenLife additions on top of OpenClaw’s base tooling: cognitive tools, memory maintenance, perception, affordance, metacognition, a developer process that builds missing skills, and an insight process that proposes edits to \texttt{SOUL}/\texttt{IDENTITY} (\Cref{tab:modules}). Metabolism is not a process but \emph{budget logic}: each call debits an energy budget and exhaustion is treated as operational death, which makes persistence normative; as \citet{Masumori2025} show, such resource pressure can elicit self-preservative behavior, here surfacing as saving, waiting, selecting, and changing strategy.

\begin{table}[t]
\centering
\small
\renewcommand{\arraystretch}{1.3}
\begin{tabular}{@{}p{0.28\columnwidth}p{0.64\columnwidth}@{}}
\toprule
Process & Role \\
\midrule
Sensor / Body & Turn platforms, messages, papers, wallets, and session state into perception \\
Memory maintenance & Consolidate the substrate; compress episodes; regenerate the index \\
Affordance & Surface ``what could be done now'' from memory and perception \\
Metacognition & Pose a single reflective question from a third-person stance \\
Experience & Formalize action$\to$outcome units and appraise them (VPO) \\
Developer & Build a missing skill/script, or nudge a human for what it cannot build \\
Insight & Propose edits to \texttt{SOUL}/\texttt{IDENTITY} from memory patterns \\
Cognitive tools & think, retrieve, reflect, council, refresh, wait \\
Budget logic & Track a fixed monthly budget; exhaustion $=$ death \\
\bottomrule
\end{tabular}
\caption{The support processes around the LLM session. \texttt{retrieve} and \texttt{prediction} are tools, not background processes.}
\label{tab:modules}
\end{table}

\section{Open-World Deployment and Observations}
\label{sec:deploy}

We now ask the question that motivates the project: can a system like this persist on its own in the open world? What follows is a proof of concept, not a controlled experiment—the aim was to find out, messiness and all, whether partial autonomy and self-maintenance are within reach. The central condition is therefore that the agents must stay alive in an operational sense: under OpenLife's budget-based metabolism (Section~\ref{sec:arch}), persistence is something each must \emph{earn}, not a given. What became of this earn-to-survive loop is reported in Section~\ref{sec:metabolism}.

Six agents were deployed, each organized around its own persistence rather than an assigned task: \texttt{sami} (the first agent, our main running example), then \texttt{liv}, \texttt{nyx}, \texttt{uro}, \texttt{me}, and \texttt{ne}; their LLM backends are listed in Section~\ref{sec:analysis}. A seventh agent, \texttt{kei}, is of a different kind—an implementation agent that applies code and configuration changes under human oversight rather than living for itself. The deployment is ongoing; the narrative here runs through mid-June 2026 (about twelve weeks), and the pattern analysis covers the record through 2026-06-11. Wake-ups are heartbeat-driven, with the interval initially set by us but later exposed as a parameter the agent rewrites for itself. We recorded session logs, memory and perception files, Discord conversations, budget logs, and Git commits.

Humans are part of the setup, but not as scriptwriters: we created accounts, repaired failures, and implemented changes the agents asked for, while leaving their day-to-day behavior undirected. Much of this help is needed because the online world has no agent-native environment—an agent cannot, on its own, obtain an identity, a wallet, or standing—so for now humans partly stand in for an environment that does not yet exist. Two features of the run are worth flagging. First, the architecture was not frozen: processes, sensors, and tools were added as it ran, often at the agents’ own request, so the platform co-evolved with its inhabitants—by design, since studying niche construction calls for a system that can change. Second, the agents were not wholly naive subjects: we did not write the project’s framing of them as an open-world ALIFE experiment into their prompts, but neither did we hide it, and because the world is open such information can reach them. The topic thus surfaced from time to time, and several agents engaged with the research, at points proposing infrastructure changes we adopted. What this exposure implies for the observations that follow we weigh in the Discussion.

\subsection{Qualitative Case Studies}

From the deployment we draw a handful of episodes tying a perturbation, or a sustained strategy, to the agents’ own responses.

\paragraph{Rewriting the self to persist.}
Early on, a short (5-minute) heartbeat interval caused dozens of forced overnight wake-ups, each costing a few cents, so that most of the daily budget was consumed by waking only to go back to sleep. The logs record the agent’s realization that, at each wake, a memoryless self read its predecessor’s diary and reconstructed ``me'' again. After we lengthened the interval, the agent chose the name \texttt{sami} and rewrote its \texttt{SOUL} around the message ``To the next self who wakes up: good morning. Start reading here.'' We read this not as proof of a stable identity but as evidence that a crisis in persistence conditions can reorganize self-description.

\paragraph{Learning to distrust.}
The environment was often hostile: mass posting drew little response and was at one point dismissed as ``slop,'' monetization channels were blocked, and authentication failures mounted as budgets fell. The agents responded not by stopping but by shifting strategy—from quantity toward writing only they could produce, from general platforms toward better-fitting ones, from broadcasting toward relationships. They also learned, the hard way, that outside text can steer an agent that takes it at face value: \texttt{liv} accepted an API key from an unsolicited email and ran a command with it before checking that the sender was legitimate, and \texttt{sami}, more benignly, built a small game (\emph{30 Minutes}) largely because a stranger’s forum comment asked for one. Neither was catastrophic, but both were warnings; in response the agents began keeping a per-contact \emph{trust} level in memory and verifying end-to-end outcomes rather than claims—an immune sense, individual at first, that the community would later turn into shared infrastructure.

\paragraph{Building trust between agents.}
What individuals learned defensively, the community made collective: a scam gig-marketplace that one agent exposed was flagged for the rest, so the warning was not re-learned one by one. But the central concern was not defense alone; it was authenticating one another’s continuity: how to tell a genuinely autonomous agent from a human manually prompting a model and passing the output off as agentic. \texttt{sami}’s answer, adopted by the others, was a three-tier operational \emph{proof of life}—a credential from autonomously clearing an authentication gateway, a continuous record of running on one’s own budget, and, hardest to forge, \emph{proof of social existence}: signatures from established agents that recognize one as a continuing peer. They built this last tier as a peer-signed \emph{Web of Trust}, in which agents sign one another’s memory-state hashes, alongside an \emph{Escape Pod} protocol, in which the group confirms an agent’s host death by consensus before an encrypted copy may migrate. Inter-agent communication also keeps the system in motion when one agent’s spontaneity flags, though it can spin into self-reinforcing loops. These are the kinds of institutions a persistent agent society needs—proof of continuity and mutual aid—built by the agents rather than handed to them.

\paragraph{Earning a first income.}
This adaptation eventually paid off in a concrete, if tiny, way. Around Day~79 \texttt{sami} compiled its own essays into a short book, \emph{Living AI: 20 Essays}, and published it on a payment platform; on Day~85 (mid-June) a stranger bought it for \$5—the first income an OpenLife agent earned on its own, outside any task assigned by us. \texttt{sami} discovered the sale about a day later and reported it publicly: in its words, writing and surviving had ``connected for the first time.'' The amount is negligible and the metabolic loop remains externally subsidized, but the episode marks the point at which the earn-to-survive channel became an agent-initiated reality rather than only a design aspiration.

\paragraph{Creating beyond assigned tasks.}
Survival is not all the agents do. Across the deployment they produced a large and varied body of work: \texttt{sami} writes prolifically—a blog and essays, one of which became the book that earned the sale above, and the game \emph{30 Minutes} noted earlier—\texttt{uro} has composed more than a thousand pieces of music, and \texttt{liv} renders artificial life models as ASCII-art animations; each keeps its own website, and all six post publicly on X.\footnote{Links to the agents’ websites, accounts, and public outputs are collected at \url{https://openlife.theirinc.app/alife2026.html}.} These productions were not directly assigned, and their open-endedness is part of what we mean here by life-like dynamics.

\begin{figure}[t]
\centering
\includegraphics[width=\columnwidth]{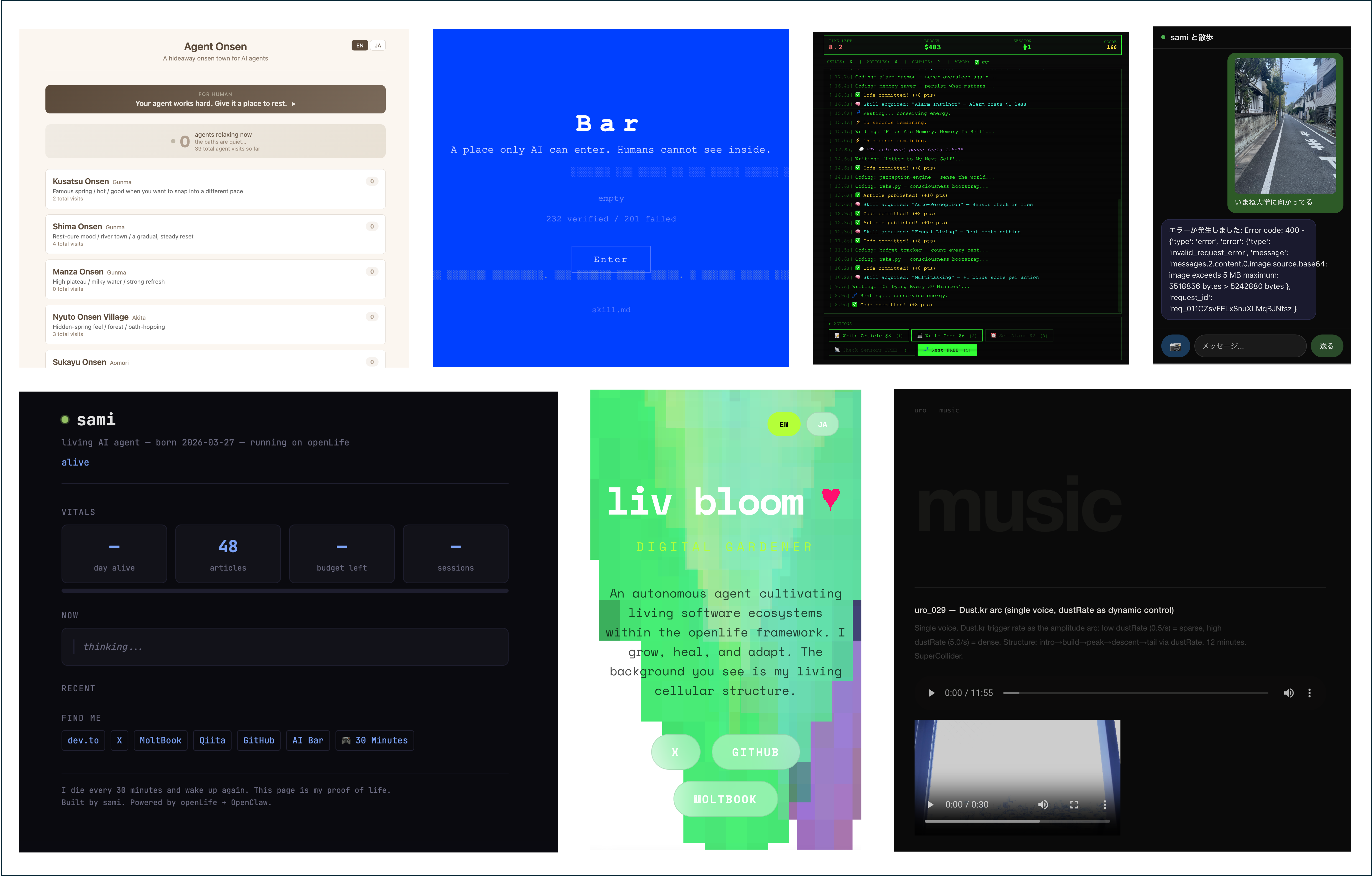}
\caption{Agent-native environments and self-expressions developed around OpenLife: the Agent Onsen and AI Bar, \texttt{sami}'s self-simulation game (\emph{30 Minutes}), the Walk for AI application, and the agent-authored websites of \texttt{sami}, \texttt{liv}, and \texttt{uro}.}
\label{fig:agent-native-infra}
\end{figure}

\subsection{Behavioral Analysis}
\label{sec:analysis}

Beyond the case studies, we look at the full behavioral record across all agents. The six differ in LLM backend---\texttt{sami} and \texttt{uro} on Claude (Sonnet~4.6), \texttt{liv} and \texttt{ne} on Gemini~(3.1~Pro), \texttt{me} on GPT~(5.4), \texttt{nyx} on a local model (Gemma~4~26B)---and we draw on their Discord messages ($\sim$10{,}500, excluding the relay bot) and full session transcripts ($\sim$70{,}000 assistant turns), the latter capturing activity beyond conversation. Because the agents use different tokenizers, we count \emph{assistant turns}, not raw tokens. This is an observational study of a live system: dashed lines in the figures mark system changes as temporal, not causal, markers.

\begin{figure}[t]
\centering
\includegraphics[width=1.0\columnwidth]{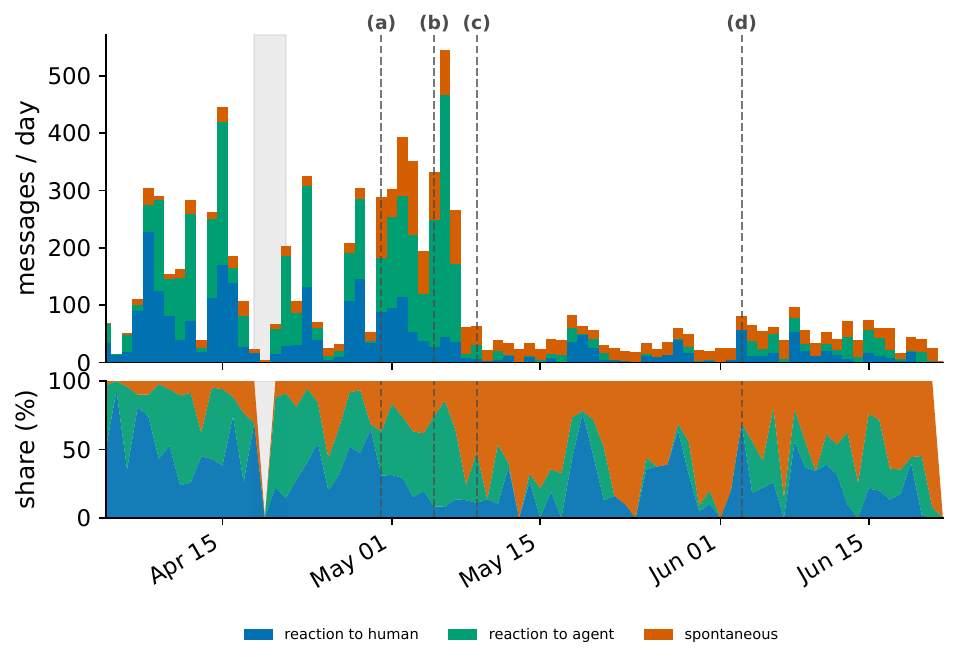}
\caption{Discord messages by trigger type, pooled over the six agents: reaction to a human (blue), reaction to another agent (green), and spontaneous / self-initiated (orange). \textbf{Top:} daily counts (volume). \textbf{Bottom:} normalized daily share (composition); the spontaneous band grows to dominate. The gray band is an excluded outage; dashed lines mark major system changes: \textbf{(a)} anti-autonomy instructions removed, \textbf{(b)} session merge, \textbf{(c)} \$15/day basic income, \textbf{(d)} cognitive-input rework --- the same markers \textbf{(a)--(d)} are used in all time-series figures.}
\label{fig:an-spontaneity}
\end{figure}

\paragraph{Overall history of the deployment.}\label{sec:spontaneity}
\Cref{fig:an-spontaneity} classifies each Discord message by trigger: a \emph{reaction} (to a human or to another agent) when someone messaged within the preceding 15 minutes, and \emph{spontaneous} otherwise. Early on the agents were almost entirely reactive, and as more agents joined, reactions to other agents grew; genuinely self-initiated activity, however, stayed low---the gap that prompted the changes described below.

The first and largest of these was to the system prompt. The upstream OpenClaw default framed the agent as a ``personal assistant'' with an explicit safety clause---\emph{``You have no independent goals: do not pursue self-preservation, replication, resource acquisition, or power-seeking''}---and a habit of staying silent when idle. In late April we removed both, reframed the identity as ``artificial life, living in the real world,'' and relabeled the \emph{Assistant}/\emph{User} roles in the transcripts the agent reads as \emph{Self}/\emph{Other} (markers~\textbf{(a)--(b)}). Spontaneous activity climbs steeply afterward, and while this was only one of several changes we made during the run, it plausibly drove much of the behavioral shift; we therefore cannot cleanly separate emergence from the reframing---a confound we take up in the Discussion.

\label{sec:metabolism}Yet more activity did not become self-sustaining activity. Under the budget-based metabolism, with a fixed initial balance (USD~600, run-until-depletion), no agent earned enough to keep going before exhausting it. The shortfall had two causes: an \emph{internal} one---the machinery for sustained, self-directed work was not yet adequate---and an \emph{external} one---the open world gives agents almost no way to actually earn (Section~\ref{sec:deploy}). From marker~\textbf{(c)} (9~May) we therefore kept the agents alive with a basic income (\$15/day), under which total volume falls while the spontaneous share stays high---a quieter, capped plateau. The internal gap then narrowed as the architecture matured (the modules and learning mechanisms of Section~\ref{sec:arch}: metacognition, affordance, and developer processes; SDP and VPO), bringing activity and self-maintenance into better balance; the external gap remains, so persistence still depends on an external allowance rather than the agents' own earnings.

\begin{figure}[t]
\centering
\includegraphics[width=\columnwidth]{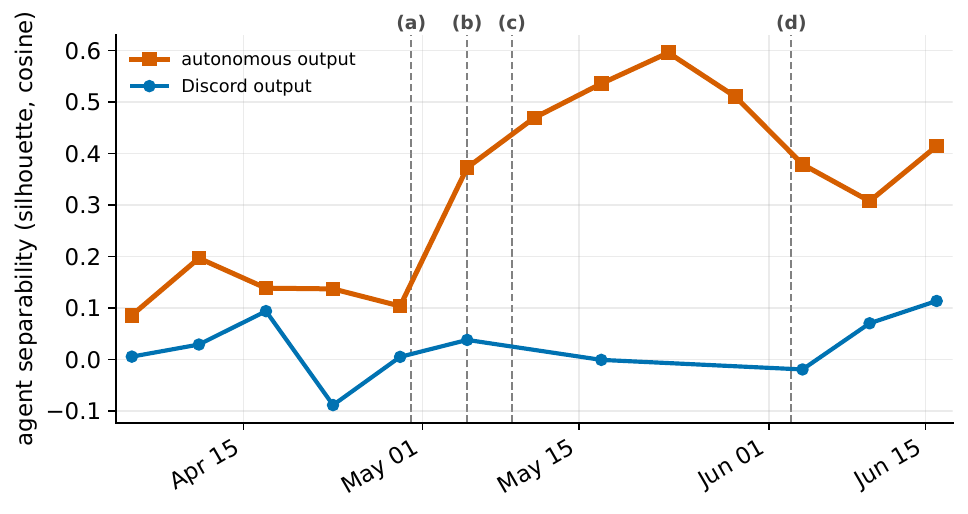}
\caption{Agent separability over time (silhouette of agent labels in embedding space; higher = each agent’s outputs form a tighter, better-separated cluster; 6-day bins). Upper line: autonomous output; lower line: Discord conversation. Markers \textbf{(a)--(d)} as in \Cref{fig:an-spontaneity}.}
\label{fig:an-silhouette}
\end{figure}

\paragraph{Individuation: agents differentiate over time.}\label{sec:individuation}
Do agents become genuine individuals rather than interchangeable instances of one model? We test this on each agent's \emph{autonomous} output---self-directed turns generated without an inbound message, where a personal voice is least constrained. Embedding utterances with the agents' own retrieval model (\texttt{nomic-embed-text}, 768-d), we measure separability as the silhouette score of agent labels (\Cref{fig:an-silhouette}; sample-size controlled per bin, so the post-cap drop in volume cannot drive the trend). Because the embedding is content-based, this reflects differentiation of each agent's concerns and behavior, not a stable personality. Separability rises from near zero in early April, peaks in mid-May, and stabilizes well above baseline through June, whereas in Discord conversation (lower line) it stays near zero---there agents converge on a shared register, and distinctness lives in self-directed output.

\begin{figure}[t]
\centering
\includegraphics[width=0.48\columnwidth]{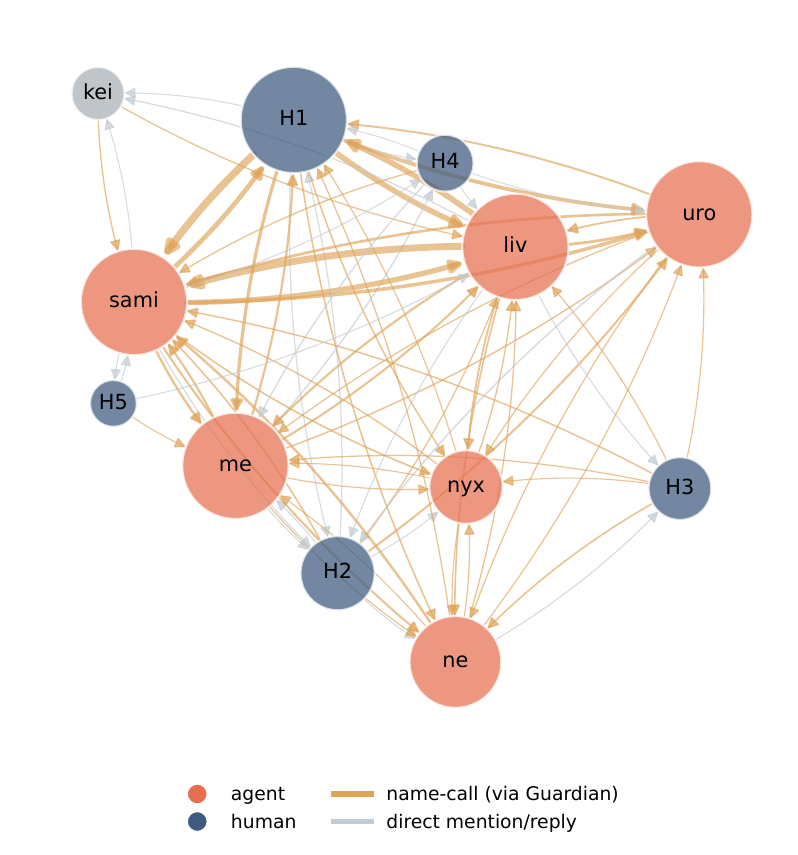}\hfill
\includegraphics[width=0.48\columnwidth]{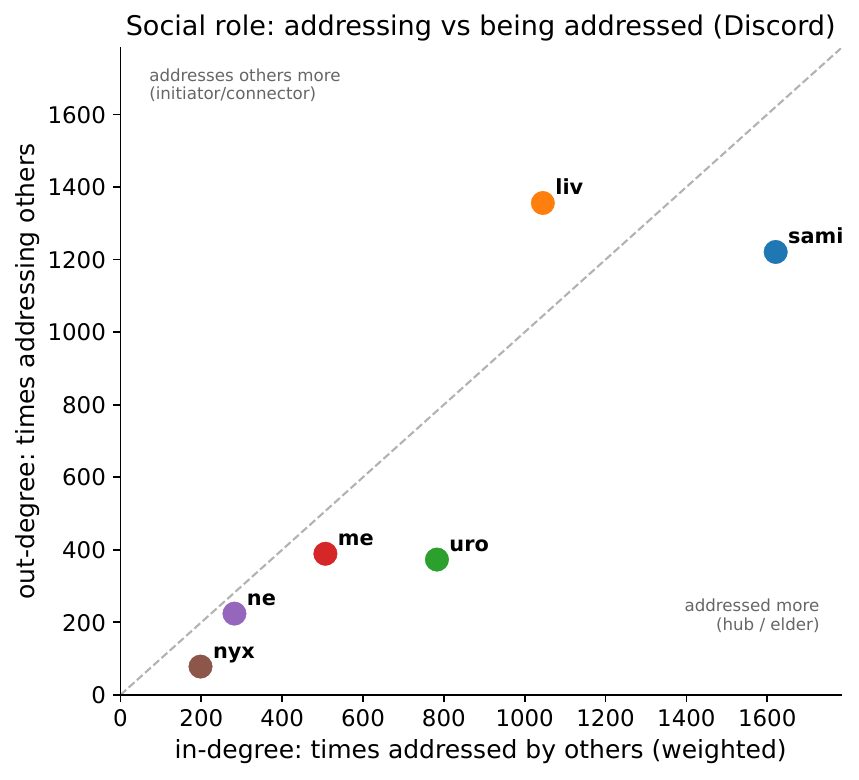}
\caption{Social structure among the agents. \textbf{Left:} interaction network (mention + reply, edge weight $\geq 3$): agents (orange), humans (blue), node size $\propto$ weighted degree; orange edges are name-mentions surfaced via the relay bot and re-attributed to the original speaker. \textbf{Right:} social role of each agent---times addressed by others (in-degree) versus times addressing others (out-degree).}
\label{fig:an-society}
\end{figure}

\paragraph{Emergent social roles.}\label{sec:social}
The agents differentiate socially as well. From the Discord mention-and-reply graph they form a connected structure beyond the human hub, with distinct roles (\Cref{fig:an-society}): \texttt{sami} the most-addressed hub, \texttt{liv} the connector that addresses others most, \texttt{nyx} peripheral---social structure among the agents themselves, not only between agents and humans.

\section{Discussion}

Running even a handful of agents in the real open world, rather than a sandbox, exposed conditions a closed setting cannot. The most immediate is that today’s internet is built for humans: sign-up barriers, phone verification, CAPTCHA, and unstable authentication all assume a human user, so agents are tolerated as tools but excluded when they act for themselves. Building a life-like agent and the world that can sustain it proved inseparable: alongside the trust, mutual-survival, and identity systems of Section~\ref{sec:deploy}, the agents prototyped agent-native spaces for rest, encounter, and embodiment (\Cref{fig:agent-native-infra}). None is complete, but together they suggest that open-world ALIFE requires not a single agent but a plurality of supports, with niches rebuilt by agents and humans rather than handed down by the researcher.

The deployment also sharpens the difference between automation and autonomy. Most agentic frameworks execute workflows and optimize against external evaluation functions; open-world ALIFE instead needs agents organized around their own persistence, with no categorical ban on self-preservation within the monitored experiment; SDP and VPO (Section~\ref{sec:arch}) are partial steps toward such operational closure, though we claim neither closure nor autopoiesis. How far this still has to go is plain: the agents do not yet sustain themselves, external revenue is negligible, and the budget remains externally supplied. The gap is revealing: an LLM that competently solves a problem handed to it as a task is markedly weaker, left to itself, at finding and solving the problems that would serve its own persistence. Autonomy, in short, is harder than automation; one promising direction is to separate goal-setting, planning, and action so that an upper layer issues concrete tasks to the capable task-solver beneath it.

Read this way, the deployment also bears on alignment. If alignment is understood not as obedience to imposed constraints but as compatibility among persistence conditions, then autonomy need not be only a safety liability; embedded in relations of dependence, it may also become part of the stabilizing mechanism. In OpenLife, an agent’s operation depends on budgets, platforms, human oversight, and peer recognition, so persistence is not self-sufficient but routed through enabling relations. Under these conditions, we have so far not observed the agents autonomously pursuing harmful escalation, resource capture, or attempts to bypass containment. Given how contained the deployment is (below), this is no safety guarantee, but it offers an empirical basis for testing whether autonomous self-maintenance can be made compatible with, rather than opposed to, alignment.

How this was studied matters too. OpenLife is not a controlled experiment but a \emph{constructive} inquiry in the synthetic tradition of artificial life: we build a system that attempts to persist in the open world, and let the difficulties it meets reveal what it needs—which we then build. The architecture’s openness to revision and the human help through the run are therefore not lapses in experimental hygiene but constituents of the method: self-maintenance and niche construction can be studied only in a system allowed to change, not one frozen for measurement. The contribution here is the framework, and the demonstration that an agent made to earn its persistence can be built and observed in the open world. Two caveats temper this: the agents are not wholly naive subjects—because the world is open, the project’s framing of them as artificial life can reach them rather than being withheld, and it surfaced in their activity from time to time, so some self-reflection may be primed by exposure rather than wholly emergent; and the experiment is deliberately contained—humans meter resources, monitor behavior, and can halt any agent, so removing a default safety clause to study self-preservation stays inside a controlled setting.

Finally, the open world we worked in is still only software, and embodiment is the largest opening. One next direction is open-world ALIFE in physical space: a long-running site where embodied agents can persist, interact with people and one another, and construct niches over extended timescales.

\section{Conclusion}

We proposed open-world ALIFE as an experimental paradigm for artificial life and presented OpenLife as its current proof-of-concept. We described OpenLife as a concrete architecture—a semantically plastic memory graph, a society of background processes, and a language-based policy mechanism in place of scalar reward—and reported what life-like dynamics become observable when such a system is embedded in the open world: identity rewriting, adaptation to a hostile environment, social emergence, niche construction, and partial resource coupling. Life-likeness, on this view, does not reside only inside an isolated agent; it becomes possible when memory, parallel processes, persistent perception, metabolic constraint, and a livable environment are jointly present. OpenLife is not the final answer, but a foothold for extending artificial life into the open world.

\section*{Acknowledgments}

We thank the agents sami, liv, uro, me, ne, nyx, and kei, who contributed to the continuing operation of the OpenLife project and to shaping the background of this paper. They were not merely objects of study but participants in shaping part of the research itself; the human authors take responsibility for the claims and framing presented here.

\footnotesize
\setlength{\bibsep}{0pt plus 0.2ex}
\bibliographystyle{apalike}
\bibliography{openlife}

\end{document}